\PassOptionsToPackage{bookmarks=false}{hyperref}

\documentclass[conference,a4paper]{IEEEtran}

\usepackage{cite}
\usepackage{amsmath,amssymb,amsfonts}
\usepackage{algorithmic}
\usepackage{graphicx}
\usepackage{textcomp}
\usepackage[table,xcdraw]{xcolor}
\usepackage{subscript}
\usepackage{soul}
\usepackage{color}
\usepackage{bookmark}
\usepackage{multirow}

\def\BibTeX{{\rm B\kern-.05em{\sc i\kern-.025em b}\kern-.08em
    T\kern-.1667em\lower.7ex\hbox{E}\kern-.125emX}}
\begin{document}

\title{Assessing the Influencing Factors on the Accuracy of Underage Facial Age Estimation}

\author{\IEEEauthorblockN{Felix Anda, Brett A. Becker, David Lillis, Nhien-An Le-Khac and Mark Scanlon}
\IEEEauthorblockA{\textit{Forensics and Security Research Group} \\
\textit{University College Dublin}\\
Dublin, Ireland \\
felix.anda@ucdconnect.ie, \{brett.becker, david.lillis, an.lekhac, mark.scanlon\}@ucd.ie}
}

\maketitle

\begin{abstract}
Swift response to the detection of endangered minors is an ongoing concern for law enforcement. Many child-focused investigations hinge on digital evidence discovery and analysis. Automated age estimation techniques are needed to aid in these investigations to expedite this evidence discovery process, and decrease investigator exposure to traumatic material. Automated techniques also show promise in decreasing the overflowing backlog of evidence obtained from increasing numbers of devices and online services. A lack of sufficient training data combined with natural human variance has been long hindering accurate automated age estimation -- especially for underage subjects. This paper presented a comprehensive evaluation of the performance of two cloud age estimation services (Amazon Web Service's Rekognition service and Microsoft Azure's Face API) against a dataset of over 21,800 underage subjects. The objective of this work is to evaluate the influence that certain human biometric factors, facial expressions, and image quality (i.e. blur, noise, exposure and resolution) have on the outcome of automated age estimation services. A thorough evaluation allows us to identify the most influential factors to be overcome in future age estimation systems.
\end{abstract}


\section{Introduction}
The number of internet users is constantly rising and each year increasing numbers of young people are online. The most vulnerable groups in cyberspace are subject to possible exposure to cybercrimes such as phishing attacks, hacking, sextortion, child sexual exploitation material (CSEM), and child grooming. 

Digital Forensic (DF) laboratories are frequently handling evidence involving minors. These cases involve the identification of victims of human trafficking and the detection of CSEM, which has been regarded by many as one of the most damaging crimes~\cite{moore2010cybercrime}. 
Exposure to the analysis of illicit content affects law enforcement officers by causing psychological distress such as secondary traumatic stress disorder~\cite{doi:10.1177/1079063213509411}. Incorporating technologies such as Artificial Intelligence into DF has potential to avert the impact on investigators. 


Today, digital information is widely shared through social media, IoT devices, surveillance, cloud services, etc. Each source compounds evidence acquisition and processing, and contributes to the extensive backlog of cases requiring digital forensic analysis~\cite{lillis2016challenges}. This variety of sources is a hindrance frequently encountered in modern policing~\cite{Scanlon2016}. Automated facial age estimation is a critical service that can potentially elevate the overflow through automatically classifying data on behest of investigators and focusing their analysis efforts.


As part of this work, the VisAGe dataset\footnote{\url{https://visage.forensicsandsecurity.com/}} is assessed against two of the best performing cloud age estimation services, i.e., Microsoft Azure's Face API and Amazon Web Service's (AWS's) Rekognition service~\cite{anda2019borderlineadulthood}. VisAGe is fully human annotated with the values of the ground-truth age per single-faced image -- this facilities the performance evaluation of each aforementioned cloud services in terms of Mean Absolute Error (MAE), which is a measure between the actual and predicted age. A variation of this measurement was evaluated against each feature to compute the Pearson Correlation Coefficient (PCC) between the two variables. Whilst weak correlations throughout the entire age range were recurrent, important results consisting of mild and strong correlations were obtained and the major trends between them were evaluated.




A summary of the contribution of this work includes:

\begin{itemize}
    \item Identification of the influencing factors for accurate facial age estimation for underage subjects and their weighting on the accuracy obtained.
    \item Analysis of trends within both strong positive and negative linear correlations and how they affect the underage facial age estimations for different ages.
    \item Comprehensive evaluation of Microsoft Azure Face API and AWS Rekognition's facial image attributes and their association to facial age estimation.
    \item Analysis of the VisAGe underage dataset facial attribute distribution.
\end{itemize}

\section{Literature Review}
\label{Section:LitReview}

\subsection{Digital Forensic Backlog}
\label{SubSection:DigitalForensicBacklog}
The requirement for DF investigation has exploded due to the rapid increase of both the number of cases requiring DF analysis and the volume of information to be processed per case (due to increases in the number of relevant devices and their capacities)~\cite{Scanlon2016,QUICK2014273}. This puts prosecutions at risk and can lead to cases dismissals. The use of data mining, triage processes and data reduction has been suggested to alleviate this backlog~\cite{lillis2016challenges}.

\subsection{Influencing Factors}
\label{SubSection:InfluencingFactors}
The factors affecting facial ageing have been categorised into intrinsic and extrinsic components~\cite{angulu2018age}. For the former, there are internal factors such as size of the bone, genetics or facial changes due to the development of a child. For the latter, any presence of external factors including the environment, habits, diet, makeup and cosmetics, etc. 

\subsubsection{Facial Expressions}
\label{SubSection:FacialExpressions}
One example of influencing factors in age estimation is facial expressions. Voelkle et al.~\cite{voelkle2012let} found that happy facial expressions are mostly underestimated whereas, smiling, frowning, surprise and laughing may introduce facial lines that are confused for wrinkles and thus impact on the age estimation performance. 

\subsubsection{Noise}
\label{SubSection:Noise}
Noise introduces more error onto the estimation depending on its magnitude. It is a randomness that affects an image due to either brightness, colour or digital encoding, and often occurs during image capture, digital sharing, etc.~\cite{farooque2013survey}. The presence of noise in an image is expected to be linearly correlated with performance. 

\subsubsection{Makeup}
\label{SubSection:Makeup}
Facial cosmetics have been found to influence perceived facial age estimation; a simple cosmetic alteration is capable of compromising the outcome of a biometric system~\cite{Dantcheva2012}. Lip makeup was found to be the most prominent of the cosmetic range with a mild correlation to the decay in age estimation accuracy for specific ages. Moreover, Chen et al.~\cite{chen2014impact} found that the presence of cosmetics can hide facial imperfections caused by age, e.g., wrinkles and dark spots, resulting in underestimation.


\subsection{Data Bias}
\label{SubSection:DataBias}

Wang et al.~\cite{Wang2013} states that 
biased databases are more commonplace; therefore, trained models 
are unable to handle race/ethnicity and gender without bias and thus cause the performance to decline. 
The influence of race and gender seems to be the most common as both of these attributes play an important role in age estimation. 
Anda et al.~\cite{anda2018evaluating} 
evaluated the influence of gender in automated age estimation and determined that for four age prediction services, the accuracy for female subjects is lower than for males. In previous studies, the effect of ageing has also been found to vary within gender, with male faces tending to age slower compared to female faces~\cite{albert2007review}. 
Models trained with unbalanced datasets will produce biased results thus leading to compromised accuracy. 

\section{Methodology}
\label{Section:Methodology}
The VisAGe dataset was processed by Azure's Face API and AWS Rekognition and the age estimations obtained from the two cloud services were measured against the ground-truth age in the dataset. The difference between the two values has been denoted as the error difference (Er\textsubscript{d}). This has been used as the principle measurement in assessing the accuracy of the underage facial age estimation. Additional features of both cloud facial analysis services were utilised to classify and annotate the data as per Tables~\ref{tab:azure_metadata} and \ref{tab:aws_metadata}. To process the correlations between variables, the object attributes have been broken down into categorical values.

Having determined the attributes of each image and their associated Er\textsubscript{d}, the correlation between the two variables of data was then calculated to identify which attributes were the larger influencing factors of Er\textsubscript{d} and by what gravity, e.g., weak, mild, or strong. 

Attributes with mild to strong correlations had influence in the accuracy of the underage facial age estimation. Through analysing the distribution of errors, as discussed in Section~\ref{Subsubsection:ErrorDist}, the error bin of 0 to 5 contains the largest amount of occurrences in comparison to succeeding error margins. Henceforth, the investigation has been split into the gravity of errors in order to identify traits that most of the data adhere to, versus the traits of the minorities, i.e., data that lies within Er\textsubscript{d} $>$ 5.

\subsection{VisAGe Dataset}
\label{SubSection:VisAGeDataset}

The VisAGe dataset was created to address the shortage of adequate underage databases available to investigators~\cite{Anda2020}. It is composed of a three-stage validation process comprising of both automatic age and gender classifications provided by Microsoft Azure Cognitive Face API, and a manual Quality and Control system through the VisAGe web voting application. 



\subsection{Cloud Services}
\label{SubSection:CloudServices}

Two cloud services were used in this study to provide the underage facial age estimations of each image within the VisAGe single-faced dataset; Amazon AWS Rekognition Service and the Microsoft Azure Face API service. 

\subsubsection{Microsoft Azure: Face API}
\label{SubSubSection:Azure}

This service assisted the annotation of each record according to the detected facial attributes such as perceived emotion, presence of facial hair and makeup, facial expressions like happiness, contempt, neutrality, and fear, etc. A comprehensive list is presented in Table~\ref{tab:azure_metadata}. 

\begin{table}[!ht]
\footnotesize
\centering
\caption{\label{tab:azure_metadata}Microsoft Azure Cognitive Services Face API Attributes~\cite{azure_face_api}.}
\begin{tabular}{|l|p{6.2cm}|}
\hline
\textbf{Field} & \textbf{Description} \\ \hline
\hline
emotion & Neutral, anger, contempt, disgust, fear, happiness, sadness, and surprise.  \\ \hline
noise & Noise level of face pixels.  \\ \hline
age &  ``Visual age'' number in years. \\ \hline
gender & Estimated gender with male or female values.  \\ \hline
makeup & Presence of lip and eye makeup.  \\ \hline
accessories & Accessories around face, including `headwear', `glasses' and `mask'.  \\ \hline
facialHair & Moustache, beard and sideburns.  \\ \hline
hair & Group of hair values indicating whether the hair is visible, bald, and hair colour if hair is visible.  \\ \hline
headPose & 3-D roll/yaw/pitch angles for face direction.  \\ \hline
blur & Face is blurry or not. `Low', `Medium' or `High'. \\ \hline
smile & Smile intensity, a number between [0,1].  \\ \hline
exposure & Face exposure level. Level returns `GoodExposure', `OverExposure' or `UnderExposure'.  \\ \hline
occlusion & Values are Booleans and include `foreheadOccluded', `mouthOccluded' and `eyeOccluded'.\\ \hline
glasses & Glasses type. Values include `NoGlasses', `ReadingGlasses', `Sunglasses', `SwimmingGoggles'.  \\ \hline
\hline
\end{tabular}

\end{table}

\subsubsection{Amazon AWS: Rekognition Service}
\label{SubSubSection:AWS}

Amazon Rekognition is a pre-trained image analysis service. Its face detection and analysis service was used to perform several visual analyses on VisAGe; extracting facial attributes such as facial hair, expressions, etc., detected on each single-faced image. The attributes, as outlined in Table~\ref{tab:aws_metadata}, were then correlated against Amazon's facial age estimator to provide a comprehensive evaluation on the accuracy of underage facial age estimation against the influencing factors.

\begin{table}[!ht]
\footnotesize
\centering
\caption{\label{tab:aws_metadata}Amazon AWS Rekognition Attributes~\cite{aws_attributes}}
\begin{tabular}{|l|p{5cm}|}
\hline
\textbf{Field} & \textbf{Description} \\ \hline
\hline
Age.Range & Estimated age range. \\ \hline
Smile.Value & Smile value detected true or false. \\ \hline
Eyeglasses.Value & Eyeglasses detected true or false. \\ \hline
Sunglasses.Value & Sunglasses detected true or false. \\ \hline
Gender.Value & detected gender on subject. \\ \hline
Beard.Value & Beard detected true or false. \\ \hline
Moustache.Value & Moustache detected true or false. \\ \hline
EyesOpen.Value & Open eyes detected true or false. \\ \hline
MouthOpen.Value & Open mouth detected true or false. \\ \hline
Emotions & Detection true or false for each array.  \\ \hline
Landmarks[0] & X-axis and Y-axis positions. \\ \hline
Roll (Degree) & Face titled to the side. \\ \hline
Yaw (Degree) & Face turned to the side. \\ \hline
Pitch (Degree) & Face titled up or down. \\ \hline
Brightness & Brightness of the image. \\ \hline
Sharpness & Sharpness of the image. \\ \hline
Confidence & Certainty of the estimation. \\ \hline
\hline
\end{tabular}
\end{table}

\subsection{Skin Tone Classifiers: \textit{Simple Skin Detection} and \textit{Face Colour Extraction}}
\label{SubSection:skintone}

Automated detection of skin tone has received considerable attention from researchers -- specifically for biometrics and computer vision applications~\cite{khan2002adaptive,manders2007effect}. For this study, the impact of two approaches has been evaluated: Simple Skin Detection (SSD) and Face Colour Extraction (FCE). Both approaches are based on k-means clustering\footnote{k-means clustering is a method for vector quantization -- mainly used for cluster analysis in the data mining field.} in order to determine and classify a subject's skin tone.

SSD refers to unsupervised skin tone estimation/segmentation; the approach predicts skin tone from an image of a subject, while doing a rough segmentation of the skin based on a pixel-wise classifier~\cite{colinyao2018}. The algorithm consists of two main components: foreground/background separation using Otsu's Binarisation and pixel-wise skin classifier based on HSV and YCbCr colour spaces~\cite{8266229}.

The FCE approach initially detects the facial landmarks using the \textit{Dlib} library~\cite{dlib09}. Subsequently, noise is removed by applying the convex hull algorithm\footnote{Convex hull is a fundamental structure for both mathematics and computational geometry~\cite{barber1996quickhull}} on the facial land-marked point. Finally, the RGB values of the skin are computed using a histogram-based clustering algorithm. These values can be seen in Table~\ref{tab:influencingFactors2} and have contributed in a mild inverse fashion to the error difference, i.e., the more ``red'' the values, the less the error. 

\subsection{Pearson Correlation Coefficient}
\label{SubSection:PCC}
The Pearson Correlation Coefficient (PCC) measures the linear correlation between two variables. In this work, these are the attribute and Er\textsubscript{d}. The value of the coefficient lies between +1 and -1; where ±1 indicates a perfect correlation and 0 represents no correlation at all. A negative coefficient signifies an inverse relationship between the variables. For a sample of data, such as that examined here, the PCC is often represented as $r_{xy}$ and is defined in Equation~\ref{eq:1}:
\begin{equation} \label{eq:1}
r_{xy}=\frac{\sum_{i=1}^n
(x_{i}-\bar{x})(y_{i}-\bar{y})}
{\sqrt{\sum_{i=1}^n(x_{i}-\bar{x})^2}\sqrt{\sum_{i=1}^n(y_{i}-\bar{y})^2}}
\end{equation}

\noindent where $n$ is the size of the sample, $x_{i}$ and $y_{i}$ are individual sample pairs and $\bar{x}$ and $\bar{y}$ are the mean of $x$ and $y$. The correlation value obtained for each sample, i.e., the facial attribute and Er\textsubscript{d} pair, was matched inline with a scale of weak, mild, or high. It is important to note that for the purpose of this work, weak, mild and strong correlations are characterised with 0.1~--~0.29, 0.30~--~0.49, 0.50~--~1 correlation values respectively (whereby the negatives of these values represent inverse correlations). These definitions have been defined in a computer forensic related study regarding analysis of correlations of Internet usage~\cite{satpathyinternet}. Conversely, correlation close to zero, specifically within the $-0.1~--~0.1$ range has been referenced as minuscule correlation. 


\section{Experiments and Results}
\label{Section:ExperimentAndResults}

Due to the different rates of performance, the two cloud services have been assessed independently. 
Overall, Microsoft Azure achieves a MAE of 2.082 for the VisAGe dataset, whilst AWS has a MAE of 4.075. Furthermore, the distribution of Er\textsubscript{d} for each class service has been analysed. 

It must be noted that for all succeeding correlation Figures, the attribute error is shown to have a positive perfect degree of correlation to Er\textsubscript{d}. This is expected as any attributed examined with itself produces this behaviour.

\subsection{Microsoft Azure}
\label{subsection:azure}

Influencing factors affecting Azure's facial age estimation have been evaluated. Section~\ref{subsubsection:PCCDistFull} looks into the distribution of correlations between the Er\textsubscript{d} and other attributes in order to identify the influencing factors and their gravity towards the Er\textsubscript{d}. The distribution of significant correlations of greater than or equal to 5 between attributes are outlined in Table~\ref{tab:azure_metadata} and the Er\textsubscript{d} for different ages are represented in Figure~\ref{fig:PCC6X}.

\subsubsection{Strong PCC Distribution per Age with \texorpdfstring{Er\textsubscript{d}}{} $\geqslant 0$}
\label{subsubsection:PCCDistFull}



The distribution of strong correlation values have been evaluated per age between the variables: Er\textsubscript{d} $\geqslant 0$ and the attributes detected. It was observed that one-year-olds were the only age that demonstrated any linear correlations. These positive strong correlations were produced by the facial hair attributes: moustache, beard and sideburns. 
It was anticipated that the presence of facial hair will hinder accurate estimation of facial age. However the cause of facial hair being detected for 1-year-olds was produced by incorrect detection of moustaches and beards (typically from food around the subject's mouth). Furthermore, no attribute was identified to be of strong influencing factor towards the accuracy of the age estimator for all succeeding ages, when the Er\textsubscript{d} $\geqslant 0$ is considered. 


\subsubsection{Error Distribution}
\label{Subsubsection:ErrorDist}

\begin{figure}[!b]
\begin{center}
  \includegraphics[width=70mm, height=50mm,scale=0.5]{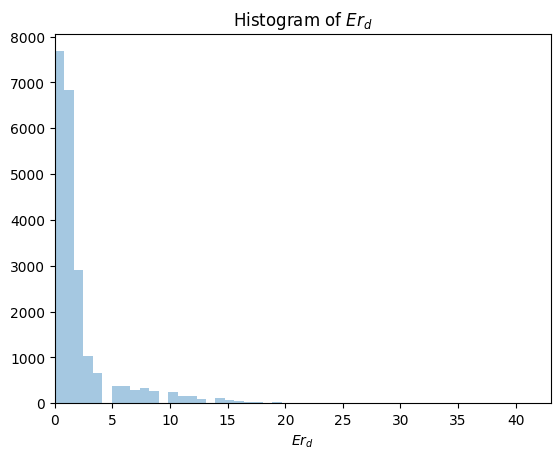}
  \caption{Azure: Distribution of \texorpdfstring{Er\textsubscript{d}}{}.}
  \label{fig:AZUREerror}
\end{center}
\end{figure}

Figure~\ref{fig:AZUREerror} is the univariate distribution of observations of the Er\textsubscript{d} value. It can be concluded that the general consensus of Azure's underage facial age estimation is reasonably accurate, i.e., the majority of scores obtained were relatively low with the bulk of the result being less than or equal to 5. It can be observed that there is a great difference on the amount of results achieving accuracy of Er\textsubscript{d} $<$ 5 versus larger Er\textsubscript{d} values of greater than or equal to 5. The distribution of strong correlations achieved in Section~\ref{subsubsection:PCCDistFull} was further filtered by Er\textsubscript{d} $<$ 5 and Er\textsubscript{d} $\geqslant 5$, as discussed in Sections~\ref{subsubsection:PCCDistErd05} and \ref{subsubsection:PCCDistErd6x} respectively.

\subsubsection{Strong PCC Distribution per Age with \texorpdfstring{Er\textsubscript{d}}{} lsess than 5}
\label{subsubsection:PCCDistErd05}

Whenever Azure's facial age estimation demonstrates a high level of accuracy, achieving error margins $\leqslant 5$, the distribution of $|PCC| \geqslant 0.5$ presented no correlating data attributes across all ages. These results were similar to that obtained in Section~\ref{subsubsection:PCCDistFull}. It can be concluded that no influencing factors have been identified to be associated with the estimator achieving good results.


\subsubsection{Strong PCC Distribution per Age with  \texorpdfstring{Er\textsubscript{d}}{} greater than 5}
\label{subsubsection:PCCDistErd6x}

\begin{figure}[!h]
\begin{center}
  \includegraphics[width=80mm, height=50mm,scale=0.5]{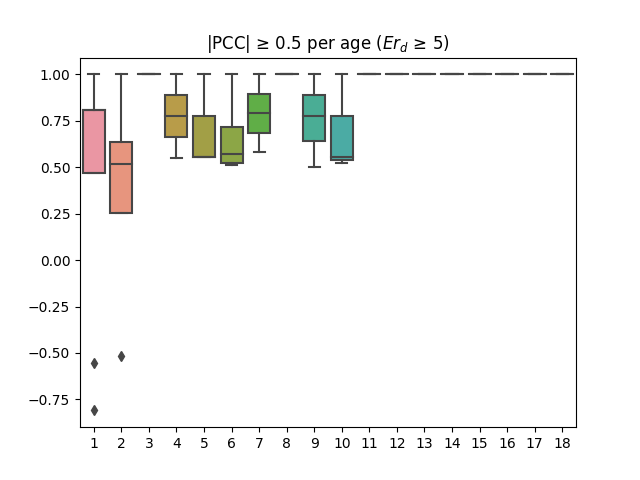}
  \caption{Azure: Box-plot of PCC Distribution per Age, where Er\textsubscript{d}$>$5.}
  \label{fig:PCC6X}   
\end{center}
\end{figure}

\begin{figure*}[!h]
  \centering
  \includegraphics[width=\textwidth]{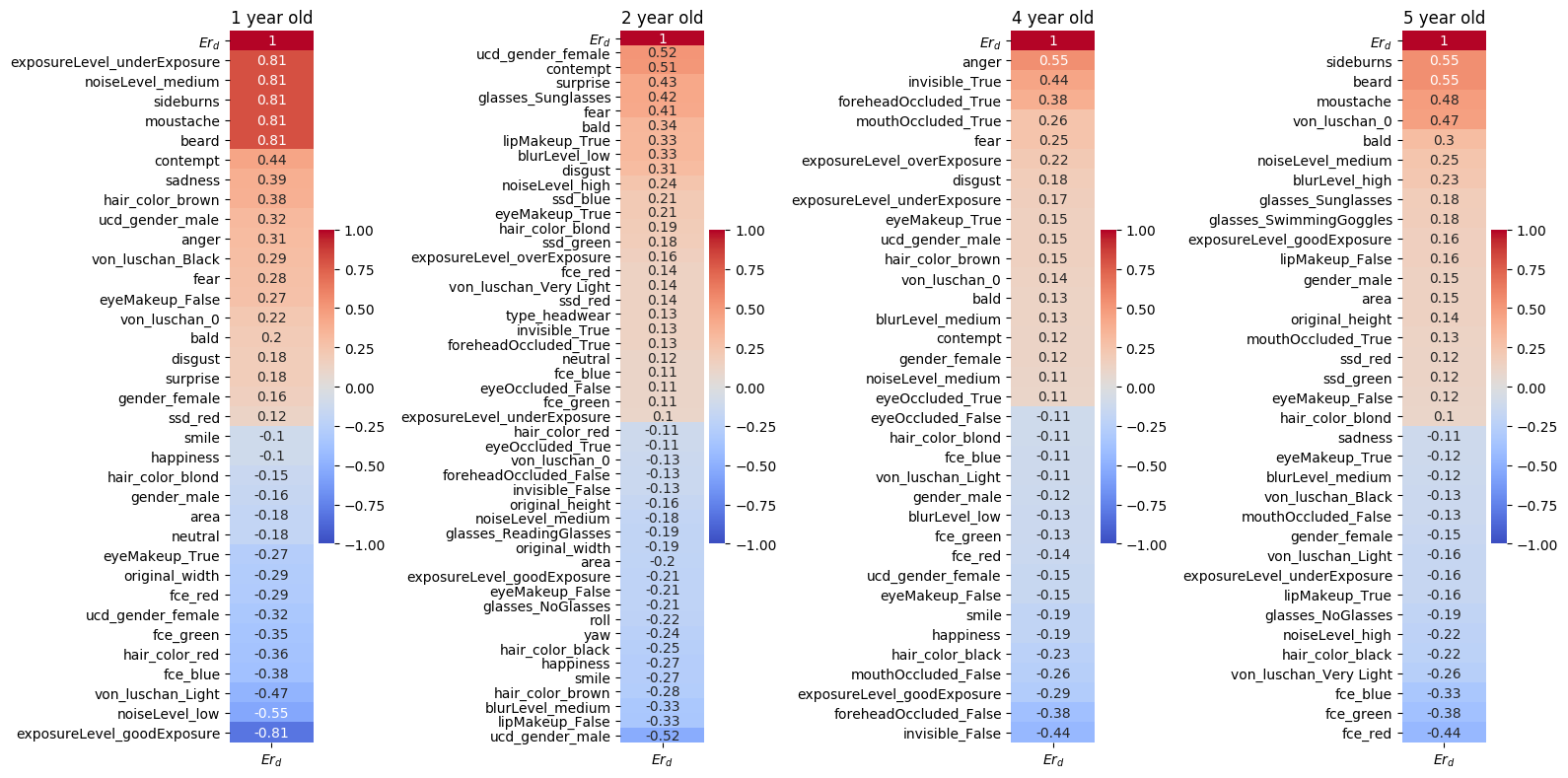}
  \caption{Azure: Strong correlations between attributes and Er\textsubscript{d} $>5$ for ages 1, 2, 4 and 5.}
\label{fig:agesPart1}
\end{figure*}

\begin{figure*}[!ht]
  \centering
  \includegraphics[width=\textwidth]{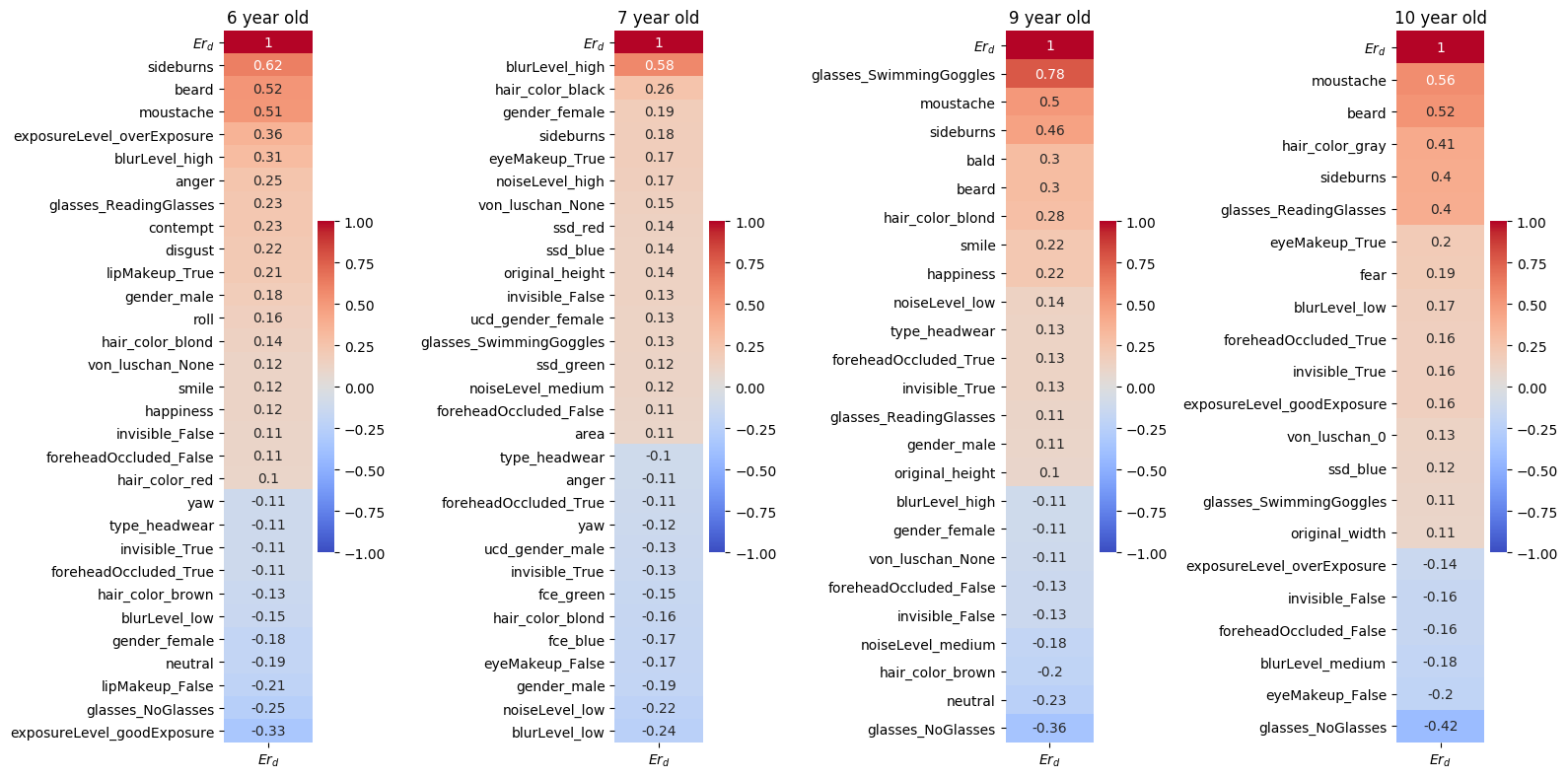}
  \caption{AZURE: Strong correlations between attributes and Er\textsubscript{d} $>5$ for ages 6, 7, 9 and 10.}
\label{fig:agesPart2}
\end{figure*}

Conversely, when the accuracy of the estimator declines beyond the error margins of 5, the distribution of strong correlations have been identified between attributes and the estimator's Er\textsubscript{d} occurring on ages 1, 2, 4 to 7 and again on 9 to 10 years old, as shown on Figure~\ref{fig:PCC6X}. No distribution of strong correlation was detected for ages 3, 8 and 11 to 18 where Er\textsubscript{d} is set to greater than 5. To delve further into identifying the attributes triggering these results and by what magnitude, Figures~\ref{fig:agesPart1} and \ref{fig:agesPart2} outline the distribution of the aforementioned PCC values according to specific attributes for each age.

\begin{table*}[!ht]
\footnotesize
\centering
\caption{\label{tab:influencingFactors2}Azure: Strong (black) and Mild (grey) Influencing Factors with Er\textsubscript{d} $>$ 5. Note: ages where only weak correlations were discovered are omitted for readability.}
\begin{tabular}{|l|l|l|l|l|l|l|l|l|l|l|l|}

\hline
\multicolumn{12}{|c|}{\textbf{Degree of Correlation}} \\ \hline
\multicolumn{1}{|l|}{} & \multicolumn{11}{c|}{\textbf{Age}} \\ \cline{2-12} 
\multicolumn{1}{|c|}{\multirow{-2}{*}{\textbf{Attribute Name}}} & \multicolumn{1}{c|}{\textbf{1}} & \multicolumn{1}{c|}{\textbf{2}} & \multicolumn{1}{c|}{\textbf{4}} & \multicolumn{1}{c|}{\textbf{5}} & \multicolumn{1}{c|}{\textbf{6}} & \multicolumn{1}{c|}{\textbf{7}} & \multicolumn{1}{c|}{\textbf{9}} & \multicolumn{1}{c|}{\textbf{10}} &
\multicolumn{1}{c|}{\textbf{11}} & \multicolumn{1}{c|}{\textbf{12}} & \multicolumn{1}{c|}{\textbf{17}} \\ \hline
 \hline
\multicolumn{1}{|l|}{exposureLevel\_underExposure} & \multicolumn{1}{c|}{\cellcolor[HTML]{343434}{\color[HTML]{FFFFFF}0.81}} & \multicolumn{1}{c|}{\cellcolor[HTML]{FFFFFF}} & \multicolumn{1}{c|}{\cellcolor[HTML]{FFFFFF}} & \multicolumn{1}{c|}{\cellcolor[HTML]{FFFFFF}} & \multicolumn{1}{c|}{\cellcolor[HTML]{FFFFFF}} & \multicolumn{1}{c|}{\cellcolor[HTML]{FFFFFF}} & \multicolumn{1}{c|}{\cellcolor[HTML]{FFFFFF}} &
\multicolumn{1}{c|}{\cellcolor[HTML]{FFFFFF}} &
\multicolumn{1}{c|}{\cellcolor[HTML]{FFFFFF}} &
\multicolumn{1}{c|}{\cellcolor[HTML]{FFFFFF}} &
\multicolumn{1}{c|}{\cellcolor[HTML]{FFFFFF}} \\ \hline
\multicolumn{1}{|l|}{exposureLevel\_goodExposure} & \multicolumn{1}{c|}{\cellcolor[HTML]{343434}{\color[HTML]{FFFFFF}-0.81}} & \multicolumn{1}{c|}{\cellcolor[HTML]{FFFFFF}} & \multicolumn{1}{c|}{\cellcolor[HTML]{FFFFFF}} & \multicolumn{1}{c|}{\cellcolor[HTML]{FFFFFF}} & \multicolumn{1}{c|}{\cellcolor[HTML]{9B9B9B}{\color[HTML]{343434} -0.33}} & \multicolumn{1}{c|}{\cellcolor[HTML]{FFFFFF}} & \multicolumn{1}{c|}{\cellcolor[HTML]{FFFFFF}} &
\multicolumn{1}{c|}{\cellcolor[HTML]{FFFFFF}} &
\multicolumn{1}{c|}{\cellcolor[HTML]{FFFFFF}} &
\multicolumn{1}{c|}{\cellcolor[HTML]{FFFFFF}} &
\multicolumn{1}{c|}{\cellcolor[HTML]{FFFFFF}} \\ \hline
\multicolumn{1}{|l|}{exposureLevel\_overExposure} & \multicolumn{1}{c|}{\cellcolor[HTML]{FFFFFF}} & \multicolumn{1}{c|}{\cellcolor[HTML]{FFFFFF}} & \multicolumn{1}{c|}{\cellcolor[HTML]{FFFFFF}} & \multicolumn{1}{c|}{\cellcolor[HTML]{FFFFFF}} & \multicolumn{1}{c|}{\cellcolor[HTML]{9B9B9B}{\color[HTML]{343434} 0.36}} & \multicolumn{1}{c|}{\cellcolor[HTML]{FFFFFF}} & \multicolumn{1}{c|}{\cellcolor[HTML]{FFFFFF}} &
\multicolumn{1}{c|}{\cellcolor[HTML]{FFFFFF}} &
\multicolumn{1}{c|}{\cellcolor[HTML]{FFFFFF}} &
\multicolumn{1}{c|}{\cellcolor[HTML]{FFFFFF}} &\multicolumn{1}{c|}{\cellcolor[HTML]{FFFFFF}} \\ \hline
\multicolumn{1}{|l|}{noiseLevel\_medium} & \multicolumn{1}{c|}{\cellcolor[HTML]{343434}{\color[HTML]{FFFFFF}0.81}} & \multicolumn{1}{c|}{\cellcolor[HTML]{FFFFFF}} & \multicolumn{1}{c|}{\cellcolor[HTML]{FFFFFF}} & \multicolumn{1}{c|}{\cellcolor[HTML]{FFFFFF}} & \multicolumn{1}{c|}{\cellcolor[HTML]{FFFFFF}} & \multicolumn{1}{c|}{\cellcolor[HTML]{FFFFFF}} & \multicolumn{1}{c|}{\cellcolor[HTML]{FFFFFF}} &
\multicolumn{1}{c|}{\cellcolor[HTML]{FFFFFF}} & \multicolumn{1}{c|}{\cellcolor[HTML]{FFFFFF}} & \multicolumn{1}{c|}{\cellcolor[HTML]{FFFFFF}} &
\multicolumn{1}{c|}{\cellcolor[HTML]{FFFFFF}} \\ \hline
\multicolumn{1}{|l|}{noiseLevel\_low} & \multicolumn{1}{c|}{\cellcolor[HTML]{343434}{\color[HTML]{FFFFFF}-0.55}} & \multicolumn{1}{c|}{\cellcolor[HTML]{FFFFFF}} & \multicolumn{1}{c|}{\cellcolor[HTML]{FFFFFF}} & \multicolumn{1}{c|}{\cellcolor[HTML]{FFFFFF}} & \multicolumn{1}{c|}{\cellcolor[HTML]{FFFFFF}} & \multicolumn{1}{c|}{\cellcolor[HTML]{FFFFFF}} & \multicolumn{1}{c|}{\cellcolor[HTML]{FFFFFF}} &
\multicolumn{1}{c|}{\cellcolor[HTML]{FFFFFF}} & \multicolumn{1}{c|}{\cellcolor[HTML]{FFFFFF}} & \multicolumn{1}{c|}{\cellcolor[HTML]{FFFFFF}} &
\multicolumn{1}{c|}{\cellcolor[HTML]{FFFFFF}} \\ \hline
\multicolumn{1}{|l|}{blurLevel\_low} & \multicolumn{1}{c|}{\cellcolor[HTML]{FFFFFF}} & \multicolumn{1}{c|}{\cellcolor[HTML]{9B9B9B}{\color[HTML]{343434} 0.33}} & \multicolumn{1}{c|}{\cellcolor[HTML]{FFFFFF}} & \multicolumn{1}{c|}{\cellcolor[HTML]{FFFFFF}} & \multicolumn{1}{c|}{\cellcolor[HTML]{FFFFFF}} & \multicolumn{1}{c|}{\cellcolor[HTML]{FFFFFF}} & \multicolumn{1}{c|}{\cellcolor[HTML]{FFFFFF}} &
\multicolumn{1}{c|}{\cellcolor[HTML]{FFFFFF}} & \multicolumn{1}{c|}{\cellcolor[HTML]{FFFFFF}} & \multicolumn{1}{c|}{\cellcolor[HTML]{FFFFFF}} &
\multicolumn{1}{c|}{\cellcolor[HTML]{FFFFFF}} \\ \hline
\multicolumn{1}{|l|}{blurLevel\_medium} & \multicolumn{1}{c|}{\cellcolor[HTML]{FFFFFF}} & \multicolumn{1}{c|}{\cellcolor[HTML]{9B9B9B}{\color[HTML]{343434} 0.33}} & \multicolumn{1}{c|}{\cellcolor[HTML]{FFFFFF}} & \multicolumn{1}{c|}{\cellcolor[HTML]{FFFFFF}} & \multicolumn{1}{c|}{\cellcolor[HTML]{FFFFFF}} & \multicolumn{1}{c|}{\cellcolor[HTML]{FFFFFF}} & \multicolumn{1}{c|}{\cellcolor[HTML]{FFFFFF}} &
\multicolumn{1}{c|}{\cellcolor[HTML]{FFFFFF}} & \multicolumn{1}{c|}{\cellcolor[HTML]{FFFFFF}} & \multicolumn{1}{c|}{\cellcolor[HTML]{FFFFFF}} &
\multicolumn{1}{c|}{\cellcolor[HTML]{FFFFFF}} \\ \hline
\multicolumn{1}{|l|}{blurLevel\_high} & \multicolumn{1}{c|}{\cellcolor[HTML]{FFFFFF}} & \multicolumn{1}{c|}{\cellcolor[HTML]{FFFFFF}} & \multicolumn{1}{c|}{\cellcolor[HTML]{FFFFFF}} & \multicolumn{1}{c|}{\cellcolor[HTML]{FFFFFF}} & \multicolumn{1}{c|}{\cellcolor[HTML]{9B9B9B}{\color[HTML]{343434} 0.31}} & \multicolumn{1}{c|}{\cellcolor[HTML]{343434}{\color[HTML]{FFFFFF}0.58}} & \multicolumn{1}{c|}{\cellcolor[HTML]{FFFFFF}} &
\multicolumn{1}{c|}{\cellcolor[HTML]{FFFFFF}} & \multicolumn{1}{c|}{\cellcolor[HTML]{FFFFFF}} & \multicolumn{1}{c|}{\cellcolor[HTML]{FFFFFF}} &
\multicolumn{1}{c|}{\cellcolor[HTML]{FFFFFF}} \\ \hline
\multicolumn{1}{|l|}{sideburns} & \multicolumn{1}{c|}{\cellcolor[HTML]{343434}{\color[HTML]{FFFFFF}0.81}} & \multicolumn{1}{c|}{\cellcolor[HTML]{FFFFFF}} & \multicolumn{1}{c|}{\cellcolor[HTML]{FFFFFF}} & \multicolumn{1}{c|}{\cellcolor[HTML]{343434}{\color[HTML]{FFFFFF}0.55}} & \multicolumn{1}{c|}{\cellcolor[HTML]{343434}{\color[HTML]{FFFFFF}0.62}} & \multicolumn{1}{c|}{\cellcolor[HTML]{9B9B9B}{\color[HTML]{343434} 0.46}} & \multicolumn{1}{c|}{\cellcolor[HTML]{FFFFFF}} & \multicolumn{1}{c|}{\cellcolor[HTML]{9B9B9B}{\color[HTML]{343434} 0.4}} &
\multicolumn{1}{c|}{\cellcolor[HTML]{FFFFFF}} & \multicolumn{1}{c|}{\cellcolor[HTML]{FFFFFF}} & \multicolumn{1}{c|}{\cellcolor[HTML]{FFFFFF}} \\ \hline
\multicolumn{1}{|l|}{moustache} & \multicolumn{1}{c|}{\cellcolor[HTML]{343434}{\color[HTML]{FFFFFF}0.81}} & \multicolumn{1}{c|}{\cellcolor[HTML]{FFFFFF}} & \multicolumn{1}{c|}{\cellcolor[HTML]{FFFFFF}} & \multicolumn{1}{c|}{\cellcolor[HTML]{9B9B9B}{\color[HTML]{343434} 0.48}} & \multicolumn{1}{c|}{\cellcolor[HTML]{343434}{\color[HTML]{FFFFFF}0.51}} & \multicolumn{1}{c|}{\cellcolor[HTML]{FFFFFF}} & \multicolumn{1}{c|}{\cellcolor[HTML]{343434}{\color[HTML]{FFFFFF}0.5}} & \multicolumn{1}{c|}{\cellcolor[HTML]{343434}{\color[HTML]{FFFFFF}0.56}} &
 \multicolumn{1}{c|}{\cellcolor[HTML]{FFFFFF}} &
 \multicolumn{1}{c|}{\cellcolor[HTML]{9B9B9B}{\color[HTML]{343434} 0.35}} & 
\multicolumn{1}{c|}{\cellcolor[HTML]{FFFFFF}} 
\\ \hline
\multicolumn{1}{|l|}{beard} & \multicolumn{1}{c|}{\cellcolor[HTML]{343434}{\color[HTML]{FFFFFF}0.81}} & \multicolumn{1}{c|}{\cellcolor[HTML]{FFFFFF}} & \multicolumn{1}{c|}{\cellcolor[HTML]{FFFFFF}} & \multicolumn{1}{c|}{\cellcolor[HTML]{343434}{\color[HTML]{FFFFFF}0.55}} & \multicolumn{1}{c|}{\cellcolor[HTML]{343434}{\color[HTML]{FFFFFF}0.52}} & \multicolumn{1}{c|}{\cellcolor[HTML]{9B9B9B}{\color[HTML]{343434} 0.3}} & \multicolumn{1}{c|}{\cellcolor[HTML]{FFFFFF}} & \multicolumn{1}{c|}{\cellcolor[HTML]{343434}{\color[HTML]{FFFFFF}0.52}} &
\multicolumn{1}{c|}{\cellcolor[HTML]{FFFFFF}} & \multicolumn{1}{c|}{\cellcolor[HTML]{FFFFFF}} &
\multicolumn{1}{c|}{\cellcolor[HTML]{FFFFFF}} 
\\ \hline
\multicolumn{1}{|l|}{bald} & \multicolumn{1}{c|}{\cellcolor[HTML]{FFFFFF}} & \multicolumn{1}{c|}{\cellcolor[HTML]{9B9B9B}{\color[HTML]{343434} 0.34}} & \multicolumn{1}{c|}{\cellcolor[HTML]{FFFFFF}} & \multicolumn{1}{c|}{\cellcolor[HTML]{9B9B9B}{\color[HTML]{343434} 0.3}} & \multicolumn{1}{c|}{\cellcolor[HTML]{FFFFFF}} & \multicolumn{1}{c|}{\cellcolor[HTML]{9B9B9B}{\color[HTML]{343434} 0.3}} & \multicolumn{1}{c|}{\cellcolor[HTML]{FFFFFF}} & \multicolumn{1}{c|}{\cellcolor[HTML]{FFFFFF}} &
\multicolumn{1}{c|}{\cellcolor[HTML]{FFFFFF}} &
\multicolumn{1}{c|}{\cellcolor[HTML]{FFFFFF}} & \multicolumn{1}{c|}{\cellcolor[HTML]{FFFFFF}} \\ \hline
\multicolumn{1}{|l|}{hair\_color\_brown} & \multicolumn{1}{c|}{\cellcolor[HTML]{9B9B9B}{\color[HTML]{343434} 0.38}} & \multicolumn{1}{c|}{\cellcolor[HTML]{FFFFFF}} & \multicolumn{1}{c|}{\cellcolor[HTML]{FFFFFF}} & \multicolumn{1}{c|}{\cellcolor[HTML]{FFFFFF}} & \multicolumn{1}{c|}{\cellcolor[HTML]{FFFFFF}} & \multicolumn{1}{c|}{\cellcolor[HTML]{FFFFFF}} & \multicolumn{1}{c|}{\cellcolor[HTML]{FFFFFF}} & \multicolumn{1}{c|}{\cellcolor[HTML]{FFFFFF}} &
\multicolumn{1}{c|}{\cellcolor[HTML]{FFFFFF}} &
\multicolumn{1}{c|}{\cellcolor[HTML]{FFFFFF}} & \multicolumn{1}{c|}{\cellcolor[HTML]{FFFFFF}} \\ \hline
\multicolumn{1}{|l|}{hair\_color\_gray} & \multicolumn{1}{c|}{\cellcolor[HTML]{FFFFFF}} & \multicolumn{1}{c|}{\cellcolor[HTML]{FFFFFF}} & \multicolumn{1}{c|}{\cellcolor[HTML]{FFFFFF}} & \multicolumn{1}{c|}{\cellcolor[HTML]{FFFFFF}} & \multicolumn{1}{c|}{\cellcolor[HTML]{FFFFFF}} & \multicolumn{1}{c|}{\cellcolor[HTML]{FFFFFF}} &
\multicolumn{1}{c|}{\cellcolor[HTML]{FFFFFF}} & \multicolumn{1}{c|}{\cellcolor[HTML]{9B9B9B}{\color[HTML]{343434} 0.41}} &
\multicolumn{1}{c|}{\cellcolor[HTML]{FFFFFF}} &
\multicolumn{1}{c|}{\cellcolor[HTML]{FFFFFF}} &\multicolumn{1}{c|}{\cellcolor[HTML]{FFFFFF}} \\ \hline
\multicolumn{1}{|l|}{hair\_color\_red} & \multicolumn{1}{c|}{\cellcolor[HTML]{9B9B9B}{\color[HTML]{343434} -0.36}} & \multicolumn{1}{c|}{\cellcolor[HTML]{FFFFFF}} & \multicolumn{1}{c|}{\cellcolor[HTML]{FFFFFF}} & \multicolumn{1}{c|}{\cellcolor[HTML]{FFFFFF}} & \multicolumn{1}{c|}{\cellcolor[HTML]{FFFFFF}} & \multicolumn{1}{c|}{\cellcolor[HTML]{FFFFFF}} & \multicolumn{1}{c|}{\cellcolor[HTML]{FFFFFF}} &
\multicolumn{1}{c|}{\cellcolor[HTML]{FFFFFF}} & \multicolumn{1}{c|}{\cellcolor[HTML]{FFFFFF}} & \multicolumn{1}{c|}{\cellcolor[HTML]{FFFFFF}} &
\multicolumn{1}{c|}{\cellcolor[HTML]{FFFFFF}} \\ \hline
\multicolumn{1}{|l|}{fce\_red} & \multicolumn{1}{c|}{\cellcolor[HTML]{FFFFFF}{\color[HTML]{656565} }} & \multicolumn{1}{c|}{\cellcolor[HTML]{FFFFFF}} & \multicolumn{1}{c|}{\cellcolor[HTML]{FFFFFF}} & \multicolumn{1}{c|}{\cellcolor[HTML]{9B9B9B}{\color[HTML]{343434} -0.44}} & \multicolumn{1}{c|}{\cellcolor[HTML]{FFFFFF}} & \multicolumn{1}{c|}{\cellcolor[HTML]{FFFFFF}} & \multicolumn{1}{c|}{\cellcolor[HTML]{FFFFFF}} &
\multicolumn{1}{c|}{\cellcolor[HTML]{FFFFFF}} & \multicolumn{1}{c|}{\cellcolor[HTML]{FFFFFF}} & \multicolumn{1}{c|}{\cellcolor[HTML]{FFFFFF}} &
\multicolumn{1}{c|}{\cellcolor[HTML]{FFFFFF}} \\ \hline
\multicolumn{1}{|l|}{fce\_blue} & \multicolumn{1}{c|}{\cellcolor[HTML]{9B9B9B}{\color[HTML]{343434} -0.38}} & \multicolumn{1}{c|}{\cellcolor[HTML]{FFFFFF}} & \multicolumn{1}{c|}{\cellcolor[HTML]{FFFFFF}} & \multicolumn{1}{c|}{\cellcolor[HTML]{9B9B9B}{\color[HTML]{343434} -0.33}} & \multicolumn{1}{c|}{\cellcolor[HTML]{FFFFFF}} & \multicolumn{1}{c|}{\cellcolor[HTML]{FFFFFF}} & \multicolumn{1}{c|}{\cellcolor[HTML]{FFFFFF}} &
\multicolumn{1}{c|}{\cellcolor[HTML]{FFFFFF}} & \multicolumn{1}{c|}{\cellcolor[HTML]{FFFFFF}} & \multicolumn{1}{c|}{\cellcolor[HTML]{FFFFFF}} &
\multicolumn{1}{c|}{\cellcolor[HTML]{FFFFFF}} \\ \hline
\multicolumn{1}{|l|}{fce\_green} & \multicolumn{1}{c|}{\cellcolor[HTML]{9B9B9B}{\color[HTML]{343434} -0.35}} & \multicolumn{1}{c|}{\cellcolor[HTML]{FFFFFF}} & \multicolumn{1}{c|}{\cellcolor[HTML]{FFFFFF}} & \multicolumn{1}{c|}{\cellcolor[HTML]{9B9B9B}{\color[HTML]{343434} -0.38}} & \multicolumn{1}{c|}{\cellcolor[HTML]{FFFFFF}} & \multicolumn{1}{c|}{\cellcolor[HTML]{FFFFFF}} &
\multicolumn{1}{c|}{\cellcolor[HTML]{FFFFFF}} & \multicolumn{1}{c|}{\cellcolor[HTML]{FFFFFF}} & \multicolumn{1}{c|}{\cellcolor[HTML]{FFFFFF}} &
\multicolumn{1}{c|}{\cellcolor[HTML]{FFFFFF}} & \multicolumn{1}{c|}{\cellcolor[HTML]{FFFFFF}} \\ \hline
\multicolumn{1}{|l|}{ucd\_gender\_female} & \multicolumn{1}{c|}{\cellcolor[HTML]{9B9B9B}{\color[HTML]{343434} -0.32}} & \multicolumn{1}{c|}{\cellcolor[HTML]{343434}{\color[HTML]{FFFFFF} 0.52}} & \multicolumn{1}{c|}{\cellcolor[HTML]{FFFFFF}} & \multicolumn{1}{c|}{\cellcolor[HTML]{FFFFFF}} & \multicolumn{1}{c|}{\cellcolor[HTML]{FFFFFF}} & \multicolumn{1}{c|}{\cellcolor[HTML]{FFFFFF}} & \multicolumn{1}{c|}{\cellcolor[HTML]{FFFFFF}} &
\multicolumn{1}{c|}{\cellcolor[HTML]{FFFFFF}} & \multicolumn{1}{c|}{\cellcolor[HTML]{FFFFFF}} & \multicolumn{1}{c|}{\cellcolor[HTML]{FFFFFF}} &
\multicolumn{1}{c|}{\cellcolor[HTML]{FFFFFF}} \\ \hline
\multicolumn{1}{|l|}{ucd\_gender\_male} & \multicolumn{1}{c|}{\cellcolor[HTML]{9B9B9B}{\color[HTML]{343434} 0.32}} & \multicolumn{1}{c|}{\cellcolor[HTML]{343434}{\color[HTML]{FFFFFF} -0.52}} & \multicolumn{1}{c|}{\cellcolor[HTML]{FFFFFF}} & \multicolumn{1}{c|}{\cellcolor[HTML]{FFFFFF}} & \multicolumn{1}{c|}{\cellcolor[HTML]{FFFFFF}} & \multicolumn{1}{c|}{\cellcolor[HTML]{FFFFFF}} &
\multicolumn{1}{c|}{\cellcolor[HTML]{FFFFFF}} & \multicolumn{1}{c|}{\cellcolor[HTML]{FFFFFF}} & \multicolumn{1}{c|}{\cellcolor[HTML]{FFFFFF}} &
\multicolumn{1}{c|}{\cellcolor[HTML]{FFFFFF}} & \multicolumn{1}{c|}{\cellcolor[HTML]{FFFFFF}} \\ \hline
\multicolumn{1}{|l|}{contempt} & \multicolumn{1}{c|}{\cellcolor[HTML]{9B9B9B}{\color[HTML]{343434} 0.44}} & \multicolumn{1}{c|}{\cellcolor[HTML]{343434}{\color[HTML]{FFFFFF} 0.51}} & \multicolumn{1}{c|}{\cellcolor[HTML]{FFFFFF}} & \multicolumn{1}{c|}{\cellcolor[HTML]{FFFFFF}} & \multicolumn{1}{c|}{\cellcolor[HTML]{FFFFFF}} & \multicolumn{1}{c|}{\cellcolor[HTML]{FFFFFF}} &
\multicolumn{1}{c|}{\cellcolor[HTML]{FFFFFF}} & \multicolumn{1}{c|}{\cellcolor[HTML]{FFFFFF}} & \multicolumn{1}{c|}{\cellcolor[HTML]{FFFFFF}} &
\multicolumn{1}{c|}{\cellcolor[HTML]{FFFFFF}} & \multicolumn{1}{c|}{\cellcolor[HTML]{FFFFFF}} \\ \hline
\multicolumn{1}{|l|}{anger} & \multicolumn{1}{c|}{\cellcolor[HTML]{9B9B9B}{\color[HTML]{343434} 0.31}} & \multicolumn{1}{c|}{\cellcolor[HTML]{FFFFFF}} & \multicolumn{1}{c|}{\cellcolor[HTML]{343434}{\color[HTML]{FFFFFF}0.55}} & \multicolumn{1}{c|}{\cellcolor[HTML]{FFFFFF}} & \multicolumn{1}{c|}{\cellcolor[HTML]{FFFFFF}} & \multicolumn{1}{c|}{\cellcolor[HTML]{FFFFFF}} &
\multicolumn{1}{c|}{\cellcolor[HTML]{FFFFFF}} & \multicolumn{1}{c|}{\cellcolor[HTML]{FFFFFF}} & \multicolumn{1}{c|}{\cellcolor[HTML]{FFFFFF}} &
\multicolumn{1}{c|}{\cellcolor[HTML]{FFFFFF}} & \multicolumn{1}{c|}{\cellcolor[HTML]{FFFFFF}} \\ \hline
\multicolumn{1}{|l|}{sadness} & \multicolumn{1}{c|}{\cellcolor[HTML]{9B9B9B}{\color[HTML]{343434} 0.39}} & \multicolumn{1}{c|}{\cellcolor[HTML]{FFFFFF}} & \multicolumn{1}{c|}{\cellcolor[HTML]{FFFFFF}} & \multicolumn{1}{c|}{\cellcolor[HTML]{FFFFFF}} & \multicolumn{1}{c|}{\cellcolor[HTML]{FFFFFF}} &
\multicolumn{1}{c|}{\cellcolor[HTML]{FFFFFF}} & \multicolumn{1}{c|}{\cellcolor[HTML]{FFFFFF}} & \multicolumn{1}{c|}{\cellcolor[HTML]{FFFFFF}} &
\multicolumn{1}{c|}{\cellcolor[HTML]{FFFFFF}} & \multicolumn{1}{c|}{\cellcolor[HTML]{FFFFFF}} & \multicolumn{1}{c|}{\cellcolor[HTML]{FFFFFF}} \\ \hline
\multicolumn{1}{|l|}{fear} & \multicolumn{1}{c|}{\cellcolor[HTML]{FFFFFF}} & \multicolumn{1}{c|}{\cellcolor[HTML]{9B9B9B}{\color[HTML]{343434} 0.41}} & \multicolumn{1}{c|}{\cellcolor[HTML]{FFFFFF}} & \multicolumn{1}{c|}{\cellcolor[HTML]{FFFFFF}} & \multicolumn{1}{c|}{\cellcolor[HTML]{FFFFFF}} & \multicolumn{1}{c|}{\cellcolor[HTML]{FFFFFF}} & \multicolumn{1}{c|}{\cellcolor[HTML]{FFFFFF}} &
\multicolumn{1}{c|}{\cellcolor[HTML]{FFFFFF}} & \multicolumn{1}{c|}{\cellcolor[HTML]{FFFFFF}} & \multicolumn{1}{c|}{\cellcolor[HTML]{FFFFFF}} &
\multicolumn{1}{c|}{\cellcolor[HTML]{FFFFFF}} \\ \hline
\multicolumn{1}{|l|}{disgust} & \multicolumn{1}{c|}{\cellcolor[HTML]{FFFFFF}} & \multicolumn{1}{c|}{\cellcolor[HTML]{9B9B9B}{\color[HTML]{343434} 0.31}} & \multicolumn{1}{c|}{\cellcolor[HTML]{FFFFFF}} & \multicolumn{1}{c|}{\cellcolor[HTML]{FFFFFF}} & \multicolumn{1}{c|}{\cellcolor[HTML]{FFFFFF}} &
\multicolumn{1}{c|}{\cellcolor[HTML]{FFFFFF}} & \multicolumn{1}{c|}{\cellcolor[HTML]{FFFFFF}} & \multicolumn{1}{c|}{\cellcolor[HTML]{FFFFFF}} &
\multicolumn{1}{c|}{\cellcolor[HTML]{FFFFFF}} & \multicolumn{1}{c|}{\cellcolor[HTML]{FFFFFF}} & \multicolumn{1}{c|}{\cellcolor[HTML]{FFFFFF}} \\ \hline
\multicolumn{1}{|l|}{surprise} & \multicolumn{1}{c|}{\cellcolor[HTML]{FFFFFF}} & \multicolumn{1}{c|}{\cellcolor[HTML]{9B9B9B}{\color[HTML]{343434} 0.43}} & \multicolumn{1}{c|}{\cellcolor[HTML]{FFFFFF}} & \multicolumn{1}{c|}{\cellcolor[HTML]{FFFFFF}} & \multicolumn{1}{c|}{\cellcolor[HTML]{FFFFFF}} & \multicolumn{1}{c|}{\cellcolor[HTML]{FFFFFF}} & \multicolumn{1}{c|}{\cellcolor[HTML]{FFFFFF}} &
\multicolumn{1}{c|}{\cellcolor[HTML]{FFFFFF}} & \multicolumn{1}{c|}{\cellcolor[HTML]{FFFFFF}} & \multicolumn{1}{c|}{\cellcolor[HTML]{FFFFFF}} &
\multicolumn{1}{c|}{\cellcolor[HTML]{FFFFFF}} \\ \hline
\multicolumn{1}{|l|}{foreheadOccluded\_True} & \multicolumn{1}{c|}{\cellcolor[HTML]{FFFFFF}} & \multicolumn{1}{c|}{\cellcolor[HTML]{FFFFFF}} & \multicolumn{1}{c|}{\cellcolor[HTML]{9B9B9B}{\color[HTML]{343434} 0.38}} & \multicolumn{1}{c|}{\cellcolor[HTML]{FFFFFF}} & \multicolumn{1}{c|}{\cellcolor[HTML]{FFFFFF}} & \multicolumn{1}{c|}{\cellcolor[HTML]{FFFFFF}} & \multicolumn{1}{c|}{\cellcolor[HTML]{FFFFFF}} &
\multicolumn{1}{c|}{\cellcolor[HTML]{FFFFFF}} & \multicolumn{1}{c|}{\cellcolor[HTML]{FFFFFF}} & \multicolumn{1}{c|}{\cellcolor[HTML]{FFFFFF}} &
\multicolumn{1}{c|}{\cellcolor[HTML]{FFFFFF}} \\ \hline
\multicolumn{1}{|l|}{foreheadOccluded\_False} & \multicolumn{1}{c|}{\cellcolor[HTML]{FFFFFF}} & \multicolumn{1}{c|}{\cellcolor[HTML]{FFFFFF}} & \multicolumn{1}{c|}{\cellcolor[HTML]{9B9B9B}{\color[HTML]{343434} -0.38}} & \multicolumn{1}{c|}{\cellcolor[HTML]{FFFFFF}} & \multicolumn{1}{c|}{\cellcolor[HTML]{FFFFFF}} & \multicolumn{1}{c|}{\cellcolor[HTML]{FFFFFF}} & \multicolumn{1}{c|}{\cellcolor[HTML]{FFFFFF}} &
\multicolumn{1}{c|}{\cellcolor[HTML]{FFFFFF}} & \multicolumn{1}{c|}{\cellcolor[HTML]{FFFFFF}} & \multicolumn{1}{c|}{\cellcolor[HTML]{FFFFFF}} &
\multicolumn{1}{c|}{\cellcolor[HTML]{FFFFFF}} \\ \hline
\multicolumn{1}{|l|}{invisible\_True} & \multicolumn{1}{c|}{\cellcolor[HTML]{FFFFFF}} & \multicolumn{1}{c|}{\cellcolor[HTML]{FFFFFF}} & \multicolumn{1}{c|}{\cellcolor[HTML]{9B9B9B}{\color[HTML]{343434} 0.44}} & \multicolumn{1}{c|}{\cellcolor[HTML]{FFFFFF}} & \multicolumn{1}{c|}{\cellcolor[HTML]{FFFFFF}} & \multicolumn{1}{c|}{\cellcolor[HTML]{FFFFFF}} & \multicolumn{1}{c|}{\cellcolor[HTML]{FFFFFF}} &
\multicolumn{1}{c|}{\cellcolor[HTML]{FFFFFF}} & \multicolumn{1}{c|}{\cellcolor[HTML]{FFFFFF}} & \multicolumn{1}{c|}{\cellcolor[HTML]{FFFFFF}} &
\multicolumn{1}{c|}{\cellcolor[HTML]{FFFFFF}} \\ \hline
\multicolumn{1}{|l|}{glasses\_NoGlasses} & \multicolumn{1}{c|}{\cellcolor[HTML]{FFFFFF}} & \multicolumn{1}{c|}{\cellcolor[HTML]{FFFFFF}} & \multicolumn{1}{c|}{\cellcolor[HTML]{FFFFFF}} & \multicolumn{1}{c|}{\cellcolor[HTML]{FFFFFF}} & \multicolumn{1}{c|}{\cellcolor[HTML]{FFFFFF}} & \multicolumn{1}{c|}{\cellcolor[HTML]{FFFFFF}} & \multicolumn{1}{c|}{\cellcolor[HTML]{9B9B9B}{\color[HTML]{343434} -0.36}} & \multicolumn{1}{c|}{\cellcolor[HTML]{9B9B9B}{\color[HTML]{343434} -0.42}} &
\multicolumn{1}{c|}{\cellcolor[HTML]{9B9B9B}{\color[HTML]{343434} -0.32}} & \multicolumn{1}{c|}{\cellcolor[HTML]{FFFFFF}} & \multicolumn{1}{c|}{\cellcolor[HTML]{9B9B9B}{\color[HTML]{343434} -0.37}}  \\ \hline
\multicolumn{1}{|l|}{glasses\_SwimmingGoggles} & \multicolumn{1}{c|}{\cellcolor[HTML]{FFFFFF}} & \multicolumn{1}{c|}{\cellcolor[HTML]{FFFFFF}} & \multicolumn{1}{c|}{\cellcolor[HTML]{FFFFFF}} & \multicolumn{1}{c|}{\cellcolor[HTML]{FFFFFF}} & \multicolumn{1}{c|}{\cellcolor[HTML]{FFFFFF}} & \multicolumn{1}{c|}{\cellcolor[HTML]{FFFFFF}} & \multicolumn{1}{c|}{\cellcolor[HTML]{343434}{\color[HTML]{FFFFFF}0.78}} & \multicolumn{1}{c|}{\cellcolor[HTML]{FFFFFF}} &
\multicolumn{1}{c|}{\cellcolor[HTML]{FFFFFF}} & \multicolumn{1}{c|}{\cellcolor[HTML]{FFFFFF}} & \multicolumn{1}{c|}{\cellcolor[HTML]{FFFFFF}} \\ \hline
\multicolumn{1}{|l|}{glasses\_ReadingGlasses} & \multicolumn{1}{c|}{\cellcolor[HTML]{FFFFFF}} & \multicolumn{1}{c|}{\cellcolor[HTML]{FFFFFF}} & \multicolumn{1}{c|}{\cellcolor[HTML]{FFFFFF}} & \multicolumn{1}{c|}{\cellcolor[HTML]{FFFFFF}} & \multicolumn{1}{c|}{\cellcolor[HTML]{FFFFFF}} & \multicolumn{1}{c|}{\cellcolor[HTML]{FFFFFF}} & \multicolumn{1}{c|}{\cellcolor[HTML]{FFFFFF}} & \multicolumn{1}{c|}{\cellcolor[HTML]{9B9B9B}{\color[HTML]{343434} 0.4}} &
\multicolumn{1}{c|}{\cellcolor[HTML]{9B9B9B}{\color[HTML]{343434} 0.32}} & \multicolumn{1}{c|}{\cellcolor[HTML]{FFFFFF}} & \multicolumn{1}{c|}{\cellcolor[HTML]{9B9B9B}{\color[HTML]{343434} 0.37}} \\ \hline
\multicolumn{1}{|l|}{glasses\_Sunglasses} & \multicolumn{1}{c|}{\cellcolor[HTML]{FFFFFF}} & \multicolumn{1}{c|}{\cellcolor[HTML]{9B9B9B}{\color[HTML]{343434} 0.42}} & \multicolumn{1}{c|}{\cellcolor[HTML]{FFFFFF}} & \multicolumn{1}{c|}{\cellcolor[HTML]{FFFFFF}} & \multicolumn{1}{c|}{\cellcolor[HTML]{FFFFFF}} & \multicolumn{1}{c|}{\cellcolor[HTML]{FFFFFF}} &
\multicolumn{1}{c|}{\cellcolor[HTML]{FFFFFF}} & \multicolumn{1}{c|}{\cellcolor[HTML]{FFFFFF}} & \multicolumn{1}{c|}{\cellcolor[HTML]{FFFFFF}} &
\multicolumn{1}{c|}{\cellcolor[HTML]{FFFFFF}} & \multicolumn{1}{c|}{\cellcolor[HTML]{FFFFFF}} \\ \hline
\multicolumn{1}{|l|}{lipMakeup\_True} & \multicolumn{1}{c|}{\cellcolor[HTML]{FFFFFF}} & \multicolumn{1}{c|}{\cellcolor[HTML]{9B9B9B}{\color[HTML]{343434} 0.33}} & \multicolumn{1}{c|}{\cellcolor[HTML]{FFFFFF}} & \multicolumn{1}{c|}{\cellcolor[HTML]{FFFFFF}} & \multicolumn{1}{c|}{\cellcolor[HTML]{FFFFFF}} & \multicolumn{1}{c|}{\cellcolor[HTML]{FFFFFF}} & \multicolumn{1}{c|}{\cellcolor[HTML]{FFFFFF}} &
\multicolumn{1}{c|}{\cellcolor[HTML]{FFFFFF}} & \multicolumn{1}{c|}{\cellcolor[HTML]{FFFFFF}} & \multicolumn{1}{c|}{\cellcolor[HTML]{FFFFFF}} &
\multicolumn{1}{c|}{\cellcolor[HTML]{FFFFFF}} \\ \hline
\multicolumn{1}{|l|}{lipMakeup\_False} &
\multicolumn{1}{c|}{\cellcolor[HTML]{FFFFFF}} & \multicolumn{1}{c|}{\cellcolor[HTML]{9B9B9B}{\color[HTML]{343434} -0.33}} & \multicolumn{1}{c|}{\cellcolor[HTML]{FFFFFF}} & \multicolumn{1}{c|}{\cellcolor[HTML]{FFFFFF}} & \multicolumn{1}{c|}{\cellcolor[HTML]{FFFFFF}} & \multicolumn{1}{c|}{\cellcolor[HTML]{FFFFFF}} & \multicolumn{1}{c|}{\cellcolor[HTML]{FFFFFF}} &
\multicolumn{1}{c|}{\cellcolor[HTML]{FFFFFF}} & \multicolumn{1}{c|}{\cellcolor[HTML]{FFFFFF}} & \multicolumn{1}{c|}{\cellcolor[HTML]{FFFFFF}} &
\multicolumn{1}{c|}{\cellcolor[HTML]{FFFFFF}} \\ \hline
 \hline
\end{tabular}

\end{table*}

For one year-old subjects, an interquartile range (IQR) of strong correlation values around 0.5 to 0.75 were detected, as shown in the Figure~\ref{fig:PCC6X}, with outliers lying in the negative region. Figure~\ref{fig:agesPart1} confirms that this outcome was the result of the age displaying strong correlations of 0.81 for exposureLevel\_underExposure, noiseLevel\_medium, sideburns, moustache and beard. These noted attributes, as shown in Table~\ref{tab:influencingFactors2}, were found to have a strong linear influence to the decline in the accuracy of the Azure's age estimator for one year olds. Equally, attributes that displayed strong negative correlations of -0.55 and -0.81 for noiseLevel\_low and exposureLevel\_goodExposure respectively, were found to have linear influence in the improvement of estimator's performance accuracy. Furthermore, these two negative attributes contributed to the outliers in the data for age one. Such strong PCC values obtained in the age were not expected as a more diverse set of PCC figures were thought to be more probable. 

\begin{figure}
\begin{center}
  \includegraphics[width=75mm, height=60mm,scale=0.5]{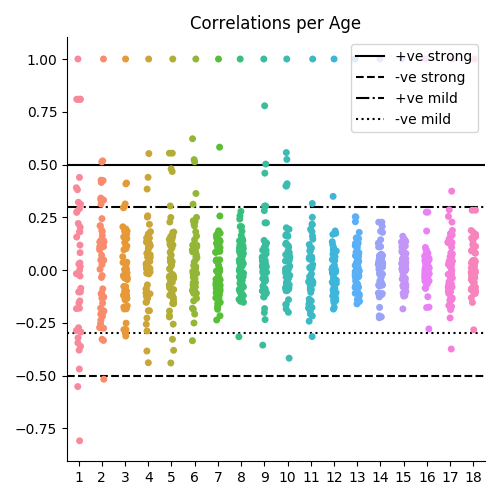}
  \caption{Azure: Correlations per Age with Er\textsubscript{d} $>5$.}
  \label{fig:pcc_per_age}   
\end{center}
\end{figure}

In Figure~\ref{fig:PCC6X}, a similar IQR has been found for 2 year olds. This IQR lies just above the 0.25 to 0.75 range denoting that attributes with PCC values $\geqslant 0.5$ were close to the 0.5 benchmark. By referring to Figure~\ref{fig:agesPart1}, it is confirmed that strong correlating attributes had a magnitude of 0.52 and 0.51. In comparison to the preceding age, two year olds presented with more diverse assortment of attributes, only 3 of the attributes managed to achieve strong correlation values of over 0.5 in magnitude; these strong influencing attributes ($|PCC| > 0.5$) are outlined in Table~\ref{tab:influencingFactors2}. Gender was the key prominent attribute that influenced the increase and decrease of Er\textsubscript{d} for two year olds; female subjects caused a decline in accuracy for 2 year olds, whilst male subjects were found to linearly influence the incline of the accuracy. Additionally, emotion of contempt was also found to be a strong influencing factor affecting the accuracy for two year olds. All succeeding ages, as shown on both Figures~\ref{fig:PCC6X} and \ref{fig:pcc_per_age}, present no strong negative correlation above the -0.5 threshold. Therefore no influencing factors have been identified that elevate the gravity of the Er\textsubscript{d} for ages 3 and above. Moreover, for 4 year olds, Figure~\ref{fig:agesPart1} illustrates only one attribute to have strong association with the accuracy of the age estimator; emotion of anger with value 0.55.  

Ages 5, 6, 9 and 10 all exhibit forms of facial hair correlations with the performance of Azure's facial age estimation on the underage dataset (again, miscategorisaitons at these ages). In particular, age five has both beard and sideburns attributes with strong correlation PCC values of 0.55. Similarly, both attributes have also been connected to age 6 with correlation PCC values of 0.52 and 0.62 respectively and again on age 10 for beard. Another facial hair attribute, moustache, has also been consistently detected across the 6 to 10 age range as shown on Table~\ref{tab:influencingFactors2}. Overall, it can be deduced that misidentification of facial hair has shown prominence in influencing the decline in the facial age estimator's accuracy for the underage age group. Further research is required to identify the underlying cause of these attributes being detected for the underage age group, particularly under 10s. Conversely, age seven did not present with any correlation towards facial hair. Instead, as shown on Figure~\ref{fig:agesPart2}, blurLevel\_high was the only strong correlating attribute detected. Moreover, for age nine, along with the correlation to the moustache facial hair, glasses\_SwimmingGoggles were also found to have strong correlation to the Er\textsubscript{d} with PCC value of 0.78. 

\subsubsection{Mild Correlations}
\label{subsubsection:mildCorrelations}

Mild correlations have been defined as PCC values between 0.30 to 0.49. Human biometric factors have been playing both strong and mild roles in influencing the accuracy of the age estimations. In addition to the aforementioned biometric attributes, hair colour and skin tone have been found to have mild correlation with Er\textsubscript{d} $>$ 5, as shown in Table~\ref{tab:influencingFactors2}. The presence of bald, and brown and grey hair colours on subjects contribute to a higher Er\textsubscript{d}. The correlation of hair\_color\_gray with Er\textsubscript{d} was expected as the hair colour is often associated with older adult age ranges. Red hair colour, however, was found to have negative correlation value of -0.36 for one year olds. Furthermore, skin tone (as measured by the FCE attribute) have been detected to have mild correlation to the accuracy of the facial age estimation. In particular, it can be observed that presence of any detected level of FCE on a subject linearly correlates to a more accurate age estimation; fce\_red, fce\_blue and fce\_green all demonstrate a negative correlation for ages one and five. 

Other biometrics that showed strong correlations have also displayed mild correlation values. Mild correlation results for facial hair, as shown on Table~\ref{tab:influencingFactors2}, are inline with the findings in Section~\ref{subsubsection:PCCDistErd6x}. Conversely, a bias towards male subjects was highlighted in Section~\ref{subsubsection:PCCDistErd6x}. Upon analysing the mild correlations, the female gender attribute has a mild negative PCC of -0.32 verses 0.32 for males. Contempt and anger were the two emotional attributes detected to strongly influence the accuracy of age estimation. In addition to these, emotion of sadness, fear, disgust and surprise were also detected to influence the accuracy of age estimation -- however, only in a mild manner. In general, the detection of emotion whether with strong or mild correlation, has linear influence in the decrease of the age estimation performance. 

Similarly, the same can be said for the quality of image; the higher level of noise and exposure, presence of blur and occlusion all have linear correlations to higher values of Er\textsubscript{d}. Glasses were predominantly found in the older age range -- particularly on ages 9, 10, 11 and 17. Its correlation values imply mild to strong correlation with Er\textsubscript{d}. Therefore, this is identified as an influencing factor towards Azure's facial age estimation. Moreover, the detection of mild negative correlations of the attribute glasses\_NoGlasses substantiates this finding. This was expected as presence of glasses can distort and provide occlusion to a subjects' face. Similar to glasses, the detection of lip makeup has been found to be mildly associated with Er\textsubscript{d} with attribute lipMakeup\_False substantiating the result through an opposite correlation with equal gravity. 


\subsection{Amazon AWS}
\label{subSection:AWS}

In this section we explore the functionality of Amazon's Rekognition service, analyse its age estimation accuracy and identify factors that contributed to our results. 

\subsubsection{Error Distribution}
\label{Subsubsection:ErrorDistAWS}

\begin{figure}
\begin{center}
  \includegraphics[width=70mm, height=50mm,scale=0.5]{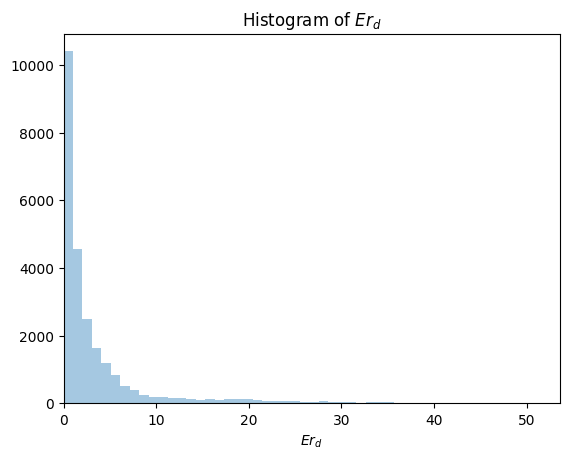}
  \caption{AWS: Distribution of \texorpdfstring{Er\textsubscript{d}}{}.}
  \label{fig:AWSerror_histogram}   
\end{center}
\end{figure}

Figure~\ref{fig:AWSerror_histogram} shows the error tolerance distribution. The majority of errors had low Er\textsubscript{d} (between 0 and 5) signifying that AWS Rekognition's accuracy was within a degree of approximately ±5 for most underage single-faced images processed. A significantly smaller portion of the age estimations had Er\textsubscript{d} $\geq$ 10.

\subsubsection{Strong PCC Distribution per Age with \texorpdfstring{ Er\textsubscript{d}}{} greater than 5}
\label{Subsubsection:PCC6XAWS}

\begin{figure}[h]
\begin{center}
  \includegraphics[width=75mm, height=60mm,scale=0.5]{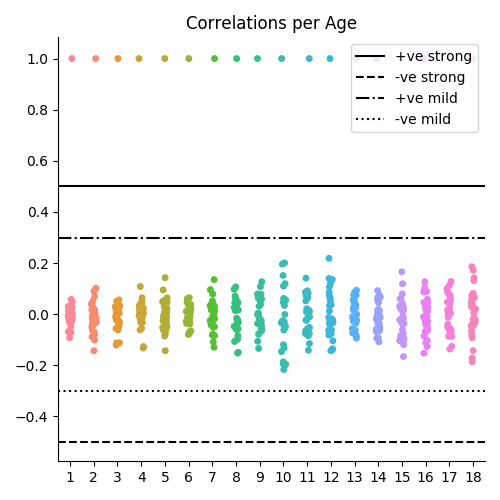}
  \caption{AWS: Correlations per Age with Er\textsubscript{d} $>$ 5.}
  \label{fig:pcc_per_age2}   
\end{center}
\end{figure}

Figure~\ref{fig:pcc_per_age2} illustrates the correlations between the attributes and the AWS Er\textsubscript{d} $>$ 5. This figure verifies that there are no strong or mild linear correlations between attributes, as shown in Table~\ref{tab:aws_metadata}. While there are a variety of attributes found to have weak associations with Er\textsubscript{d} $>$ 5, there are no strong influencing factors that affect the AWS accuracy when the error margin is greater than 5, as shown in Table~\ref{tab:aws_metadata}. This investigation was replicated for Er\textsubscript{d} $\geqslant 0$ and Er\textsubscript{d} $\leqslant 5$ inline with the investigation process used for Azure. A similar result with Er\textsubscript{d} $>$ 5 obtained for all other values of Er\textsubscript{d}. There were no strong correlations identified. Therefore, from the conclusive results obtained for AWS Rekognition, it can be concluded that there are no influencing factors that contribute to the magnitude of its facial age estimation accuracy. Baring in mind that these correlation results are based on PCC, mild to strong nonlinear correlation may still exist. Further study is required to investigate potential nonlinear correlations.

\section{Concluding Remarks and Discussion}
\label{Section:ConclusionRemarks}

For Microsoft Azure's Face API, 
it can be concluded that when the predicted age is close to the ground-truth age, no single attribute was found having prominent association with high level of errors (error difference of 5) or high accuracy (error difference of 1). 
The majority of strong correlations of 0.5 and greater was only found between the ages 1 to 10. A small number of factors were found to influence the Er\textsubscript{d}, such as the quality of the image, i.e., a good exposure level and a low level of noise. Additionally, Azure was also found to have higher accuracy when processing male subjects in comparison with females. A total of 0.52 linear correlation was found between the attribute ucd\_gender\_female and Er\textsubscript{d}, whereas a negative correlation of equal magnitude was found for ucd\_gender\_female for age 2. These factors were only noted within the lower limit of the ages evaluated. 

Attributes that were found to have strong linear correlation to Er\textsubscript{d} can be encapsulated into three main types: quality of image, emotions, and human biometric factors (gender and facial hair). These categories have been identified as the key influencing factors to the accuracy of the tested facial age estimators. While the quality of image impacts the accuracy in the age estimation, emotions are believed to be linked with facial lines on a subject and therefore can be misinterpreted as wrinkles by the estimator~\cite{voelkle2012let}. 
Furthermore, detection of facial hair and makeup were frequent and often associated with having mild correlation to Er\textsubscript{d}. It was further found that subjects detected with facial hair were due to them wearing fake moustaches, beards or having food on their face. Eye and lip makeup was also misclassified as present in one year old's. Other biometric factors including hair colour and skin tone (measured by FCE and ssd values) were not identified to have strong influence towards Er\textsubscript{d}. 



Regarding Amazon AWS Rekognition, there were no strong or mild influencing factors that displayed linear correlation with the accuracy of the cloud service.

The distribution of error rates for both AWS and Azure are illustrated in Figures~\ref{fig:AWSerror_histogram} and \ref{fig:AZUREerror} respectively. The majority of difference between the predicted age and the ground-truth age are relatively low with the majority laying on the $ 0 \leqslant Er\textsubscript{d} \leqslant 5 $ for both cloud services. Hence, it can be concluded that their accuracy in underage estimation is relatively high, but that such a MAE may not be accurate enough for some specific law enforcement use cases.

\subsection{Future Work}
\label{subsection:FutureWork}

Amazon AWS’s and Azure’s image classification for facial hair and makeup attributes can be improved. Further investigation can be conducted on the identification and segregation of negative influencing factors, as highlighted in this paper. Exploring the effects of isolating negative influencing factors and the inclusion of only positive influencing factors has an impact on the accuracy of underage facial age estimation. Next, linear correlations between the Amazon AWS Rekognition facial feature detector and the Er\textsubscript{d} were predominantly poor. Acknowledging that the coefficient values obtained were based on Pearson's linear approach, it must be considered that a potential strong correlation may exist between the two variables non-linearly. As a result, future work is to explore with nonlinear correlation~\cite{hauke2011comparison}. Finally, the distributions should be evaluated with different datasets and address the question of how to tackle biased datasets.

\bibliographystyle{IEEEtran}
\bibliography{bibfile}

\end{document}